\documentclass[a4paper,10pt,twocolumn]{article}
\usepackage[english]{babel}
\usepackage[utf8]{inputenc}
\usepackage[T1]{fontenc}

\usepackage[top=1.5cm, left=1.5cm, right=1.5cm, bottom=1.5cm]{geometry}

\renewenvironment{abstract}{\bf\small {\em\ Abstract---}}{}

\usepackage{amsfonts,amssymb,amsmath,amsthm}
\usepackage{subfigure}
\usepackage{graphicx}
\usepackage[footnotesize]{caption}

\usepackage{sectsty}
\sectionfont{\large}
\subsectionfont{\normalsize}
\usepackage[compact]{titlesec}

\title{From Adaptive Kernel Density Estimation to Sparse Mixture Models}

\author{Colas~Schretter$^{1,3}$, Jianyong~Sun$^{2}$ and Peter~Schelkens$^{1,3}$\\
\footnotesize $^1$Vrije Universiteit Brussel (VUB), Dept. of Electronics and Informatics (ETRO), B- 1050 Brussels, Belgium.\\
\footnotesize $^2$School of Mathematics and Statistics, Xi'an Jiaotong University, Xi'an 710049, China.\\
\footnotesize $^3$imec, Kapeldreef 75, B-3001 Leuven, Belgium.
\footnote{This work received funding from the European Research Council under the EU FP7/2007-2013 / ERC Grant Agreement Nr.~617779 (INTERFERE).}}

\date{\empty} 

\renewcommand\footnotemark{}

\begin{document}

\maketitle

\begin{abstract}
We introduce a balloon estimator in a generalized expectation-maximization method for estimating all parameters of a Gaussian mixture model given one data sample per mixture component. Instead of limiting explicitly the model size, this regularization strategy yields low-complexity sparse models where the number of effective mixture components reduces with an increase of a smoothing probability parameter $\mathbf{P>0}$. This semi-parametric method bridges from non-parametric adaptive kernel density estimation (KDE) to parametric ordinary least-squares when $\mathbf{P=1}$. Experiments show that simpler sparse mixture models retain the level of details present in the adaptive KDE solution.
\end{abstract}

\section{Introduction}
In bivariate adaptive kernel density estimation (KDE), we wish to solve the problem of estimating all individual variances-covariance matrices $\{\Sigma_1,\dots,\Sigma_N\} \in \mathbb{R}^{2 \times 2}$ associated to $N$ input point samples $\{x_1,\dots,x_N\} \in \mathbb{R}^2$ such that the underlying continuous density function $f(x)$ is represented by a finite Gaussian mixture model (GMM) with $M=N$ components:
\begin{align}
f(x)=\sum_{m=1}^M \pi_m \ \mathcal{N}(x|\mu_m,\Sigma_m) \quad\text{with}\quad \sum_{m=1}^M \pi_m = 1 \ , \nonumber
\end{align}
where $\mathcal{N}$ is the multivariate Gaussian probability density
\begin{align}
\mathcal{N}(x|\mu,\Sigma)=\frac{1}{2\pi\sqrt{|\Sigma|}}\exp\left[-\frac{1}{2}(x-\mu)^\top \Sigma^{-1} (x-\mu)\right]. \nonumber
\end{align}

Non-parametric estimation approaches like KDE constrain the means of mixture components to observed point samples and assumes uniform prior probabilities such that
\begin{align}
\mu_m = x_m \quad\text{and}\quad \pi_m=\frac{1}{N} \quad\text{for each}\quad m \in \left[1,\dots,N\right]. \nonumber
\end{align}
If component's means are allowed to shift, we enter the realm of semi-parametric density estimation: A highly underdetermined problem that is classically regularized by enforcing a modest model size $M<<N$ and estimating jointly all parameters with the maximum-likelihood (ML) framework, leading to simple update rules in an expectation-maximization (EM) algorithm.

Semi-parametric estimation faces two unsolved problems that are (lightly) avoided by the non-parametric framework: How to choose the model size? How to initialize parameters of mixture components? In answer, this work proposes to start from a full mixture model with $M=N$, initialized to the trivial ML solution where components fits data points with near-singular variances-covariance matrices. A regularization strategy uses a balloon estimator \cite{Sain2002} on the current density estimate for convolving locally and adaptively the data. After convergence, a sparse lower-complexity mixture model emerges. 

\begin{figure*}
\centering
\includegraphics[width=0.195\textwidth]{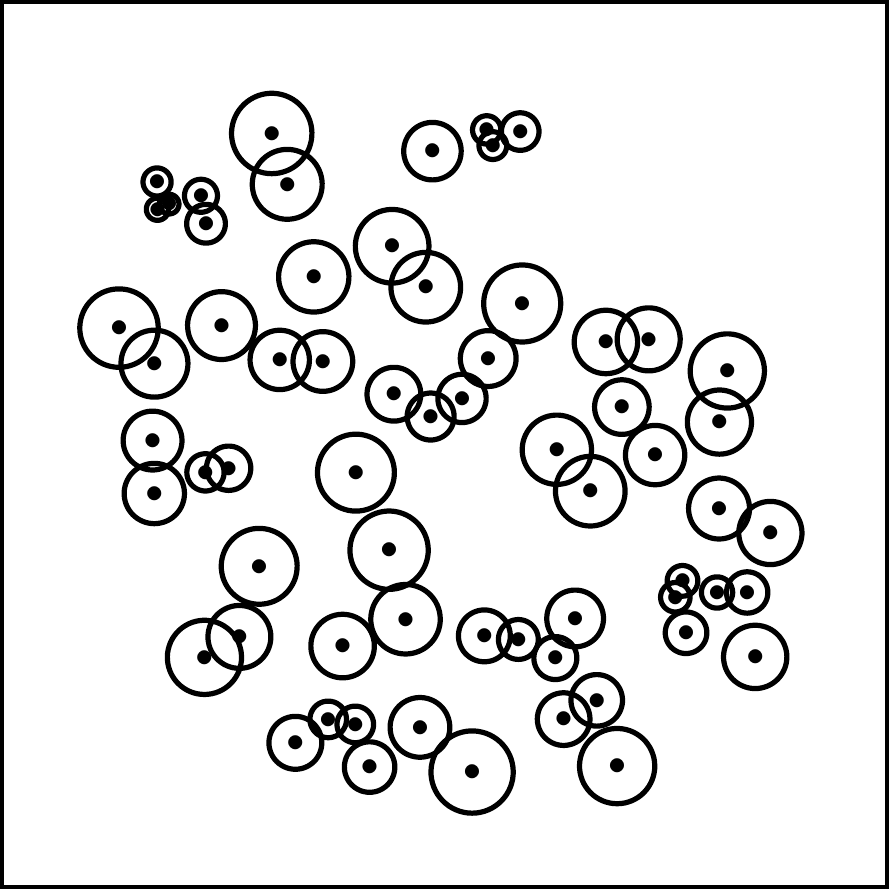}\hfill{}%
\includegraphics[width=0.195\textwidth]{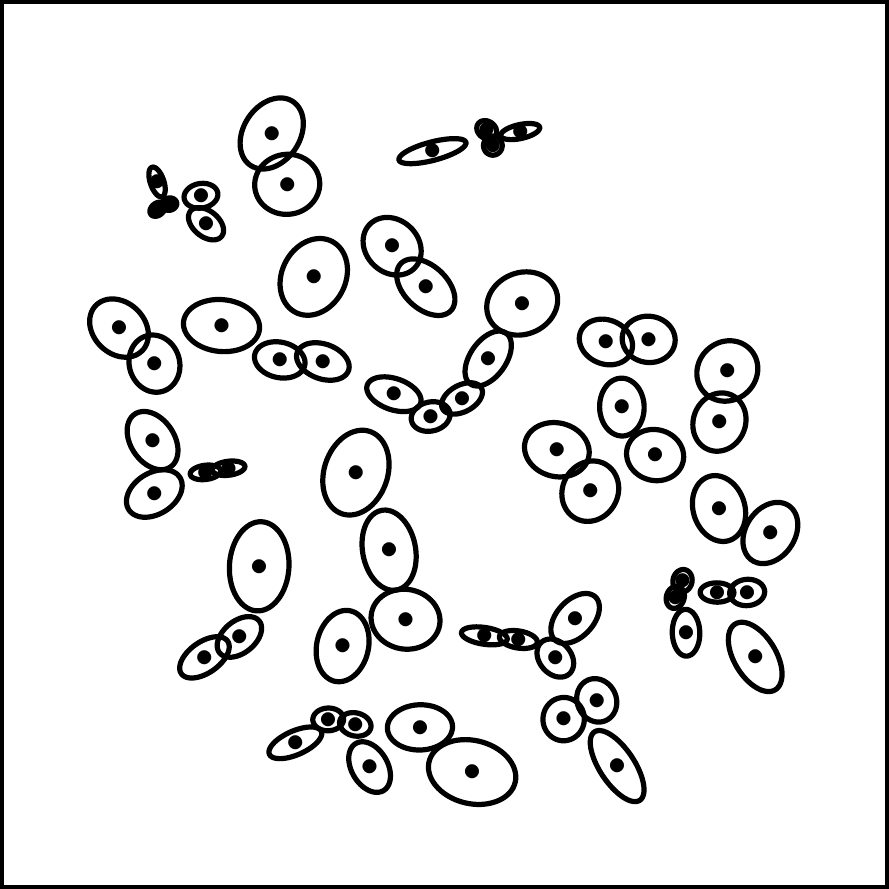}\hfill{}%
\includegraphics[width=0.195\textwidth]{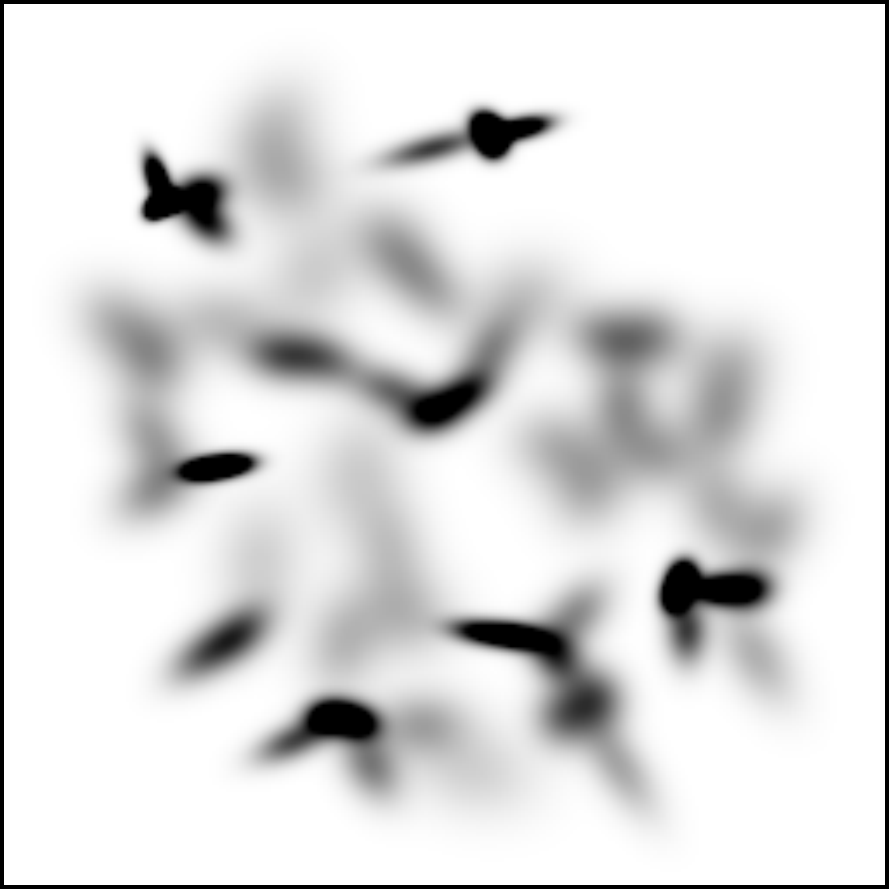}\hfill{}%
\includegraphics[width=0.195\textwidth]{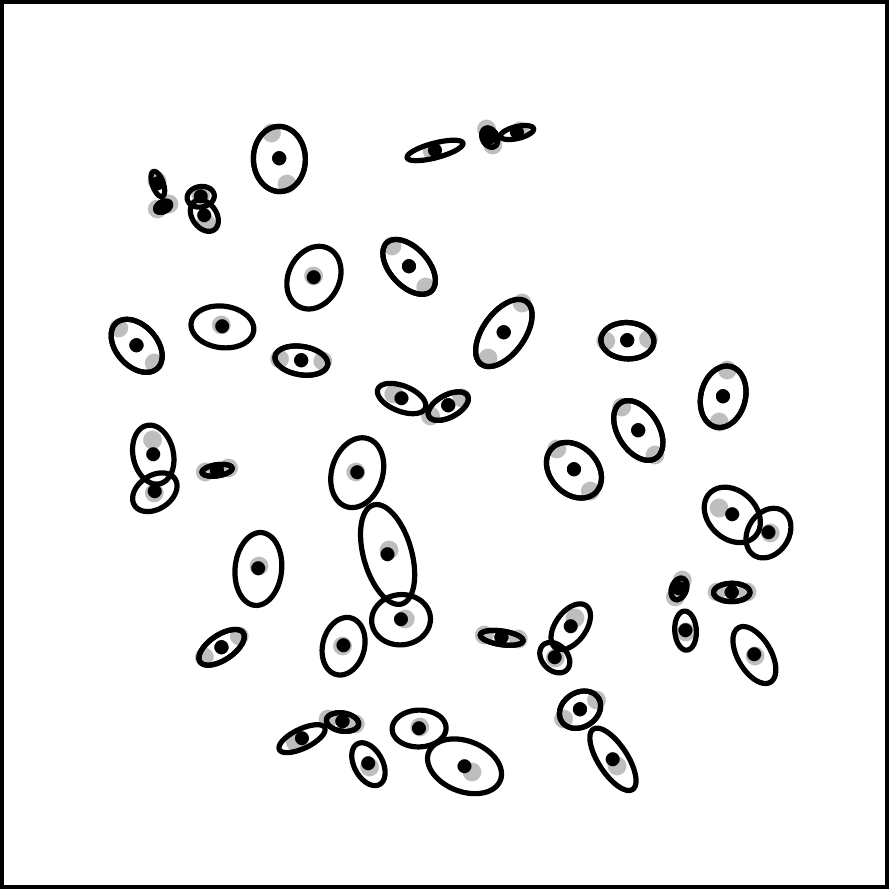}\hfill{}%
\includegraphics[width=0.195\textwidth]{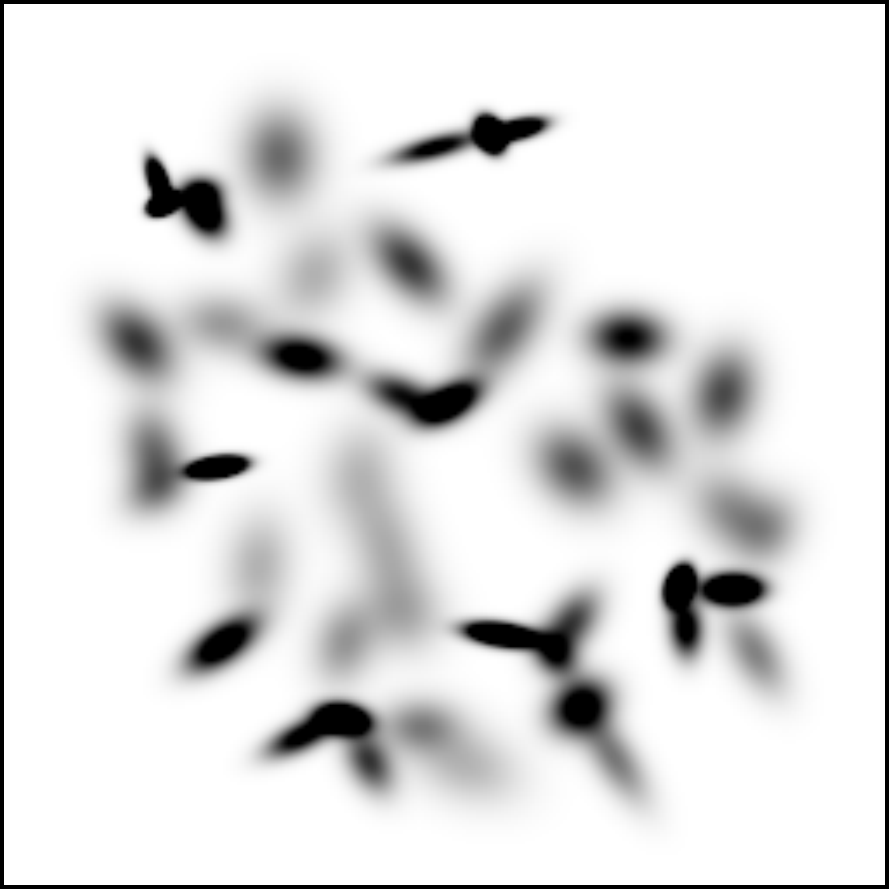}%

\vspace{0.8mm}
\includegraphics[width=0.195\textwidth]{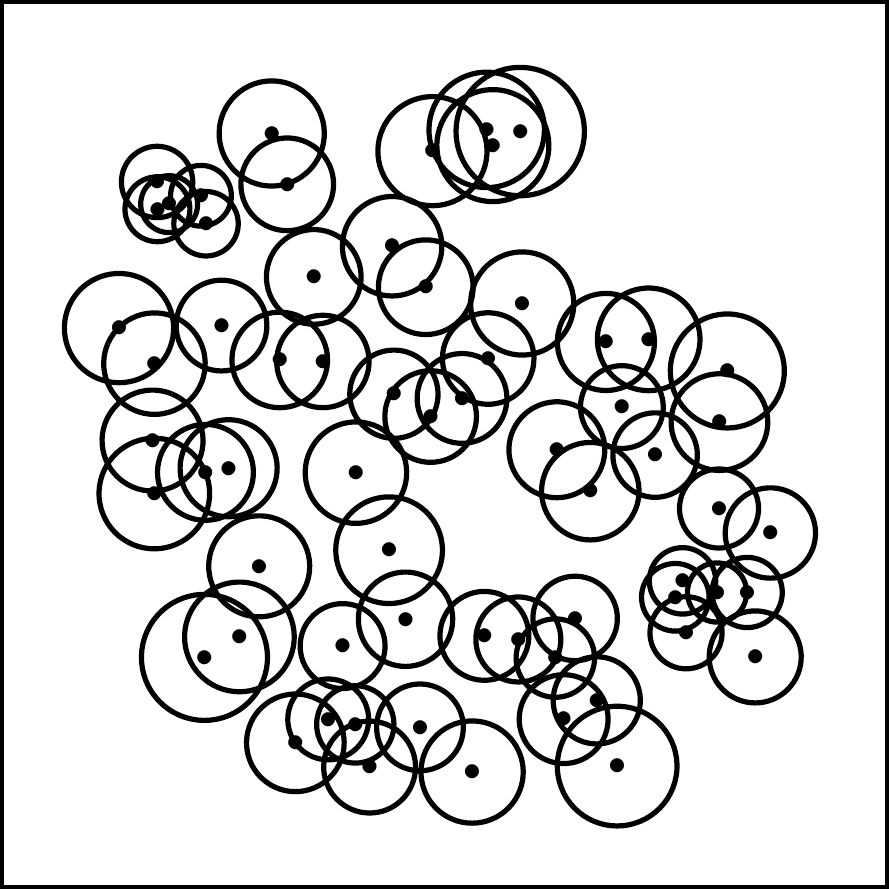}\hfill{}%
\includegraphics[width=0.195\textwidth]{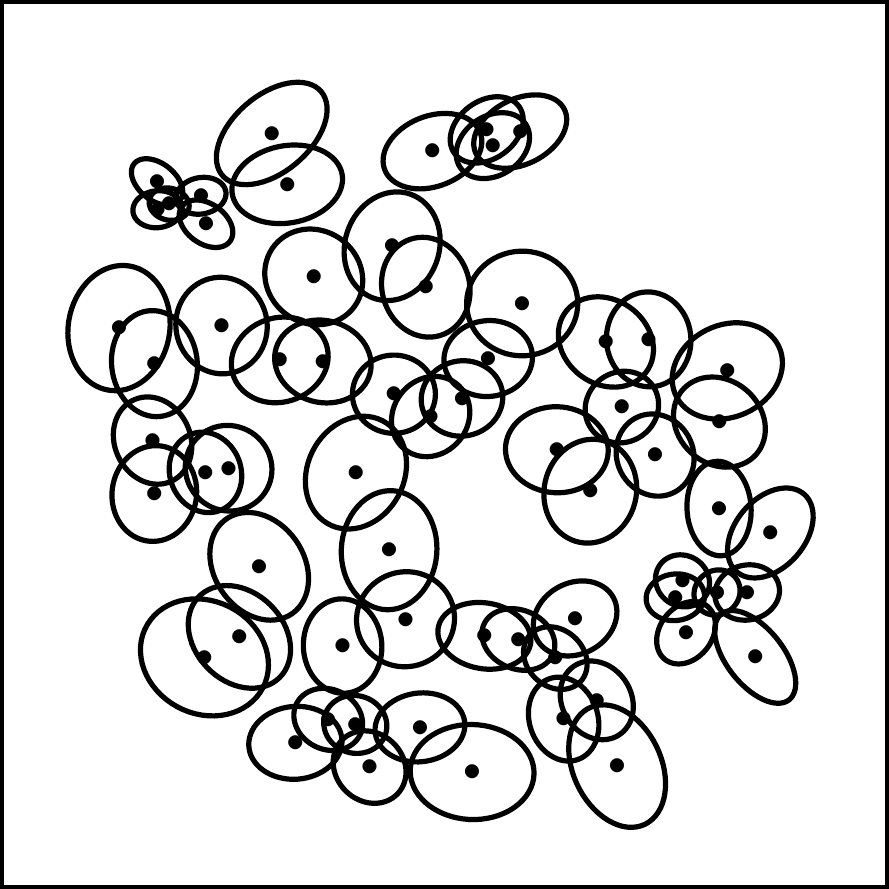}\hfill{}%
\includegraphics[width=0.195\textwidth]{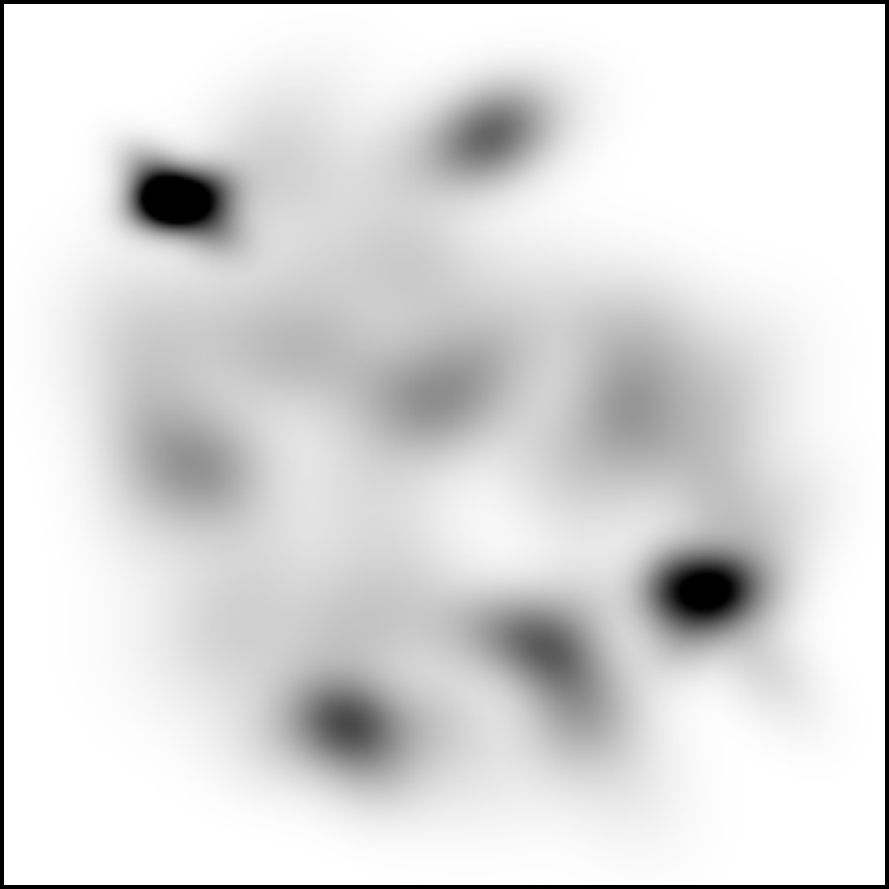}\hfill{}%
\includegraphics[width=0.195\textwidth]{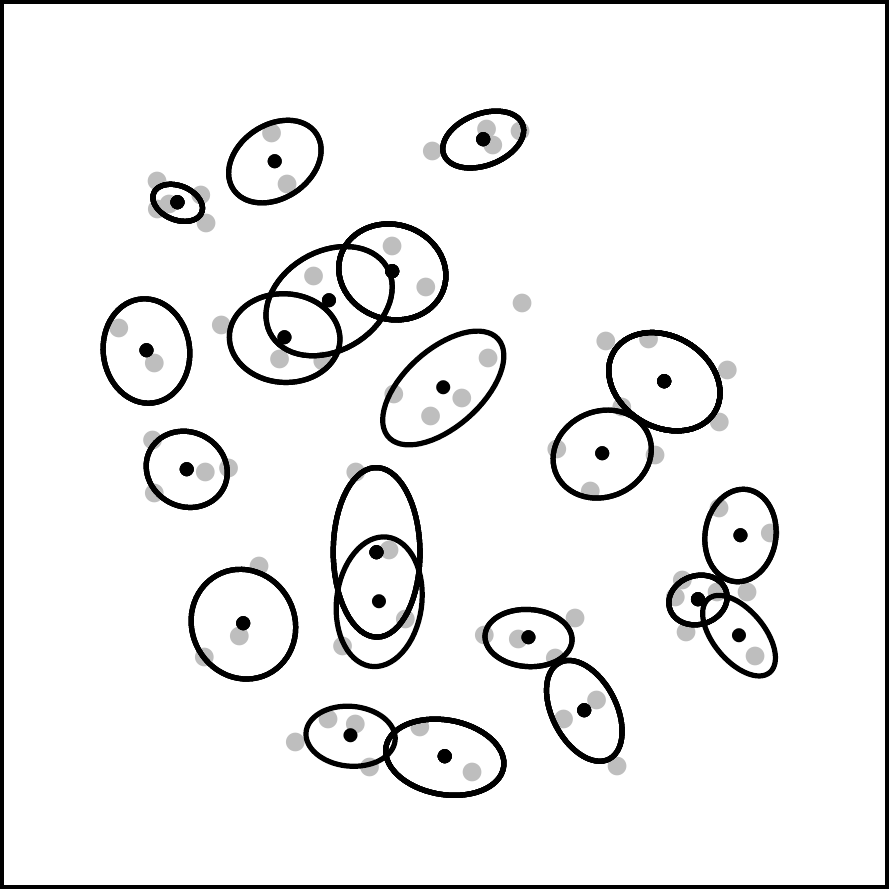}\hfill{}%
\includegraphics[width=0.195\textwidth]{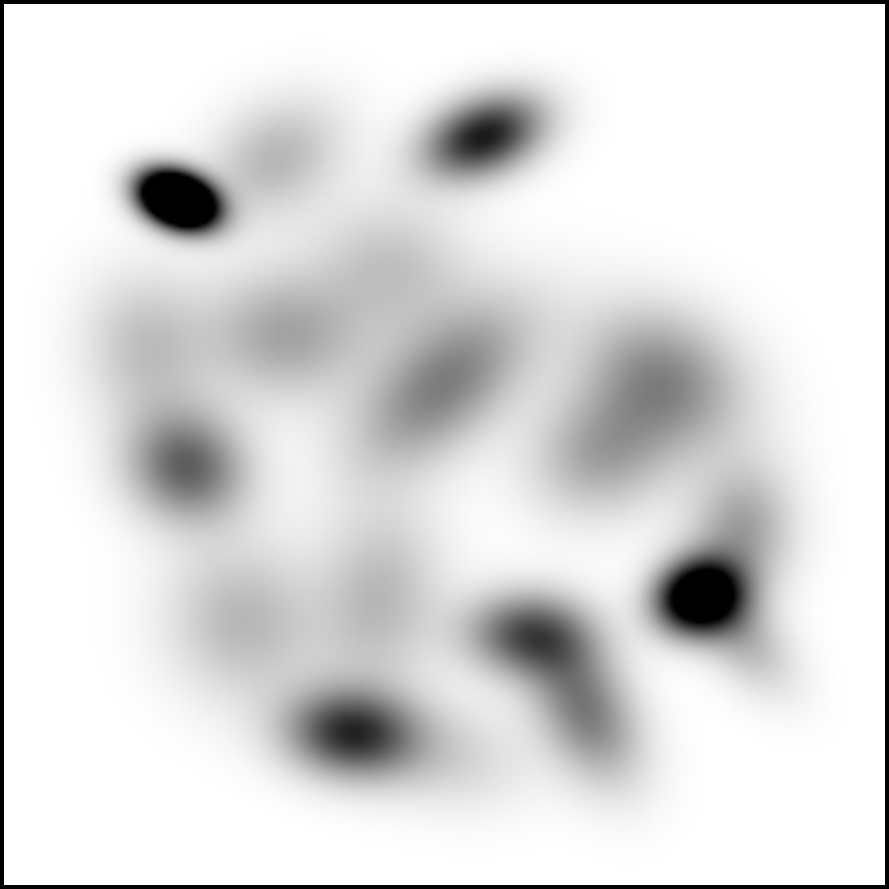}%

\makebox[0.195\textwidth]{\small Samples and balloons}\hfill{}%
\makebox[0.195\textwidth]{\small Multivariate kernels}\hfill{}%
\makebox[0.195\textwidth]{\small Adaptive KDE density}\hfill{}%
\makebox[0.195\textwidth]{\small Mixture components}\hfill{}%
\makebox[0.195\textwidth]{\small Regularized GMM density}%
\caption{\label{fig:experiments}\small Density estimates from $N=64$ randomly distributed samples, with prior probability $P=1/N$ (top) and $P=2/N$ (bottom).}
\end{figure*}

\section{Method}
The initial ML solution is attributing a very small mixture component to each data sample and fails at generalizing in regions where no sample is collected. Regularization is introduced for accounting the free space in-between samples by iteratively updating an isotropic balloon estimate per data element. A multivariate data-driven regularizing kernel is then computed analytically. Each iteration of a generalized EM algorithm updates jointly all parameters, given sample's positions and their estimated kernel representing their spatial "territory". This alternating optimization of data-space ballooning and solution-space EM converges to a stationary point. The only controllable parameter for smoothing is the prior probability $P \in (0,1]$.

\subsection{Balloon estimator}
Given a density function $f$ and any position $x \in \mathbb{R}^2$, a root finding numerical method can search for the unique variance $\sigma^2$ such that the integrated product of the density with a corresponding peak-normalized anisotropic multivariate kernel 
\begin{align}
K(r|x,R)=\exp\left[-\frac{1}{2}(r-x)^\top R^{-1} (r-x)\right], \nonumber 
\end{align}
should be equal to the given regularization parameter $P$.

For each isotropic balloon of variance $S_n=\sigma_n^2 I \in \mathbb{R}^{2 \times 2}$ centered at $x_n$, the kernel matrix $R_n$ is built by computing the ML fit of the product of the current density function with the local soft spatial "territory" covered by the balloon estimate $S_n$.

The integral of products can be evaluated in closed form since the density $f$ is expressed by a GMM and we have
\begin{align}
P(x_n|S_n)&=\sum_{m=1}^M \pi_m \int \mathcal{N}(r|\mu_m,\Sigma_m) \ K(r|x_n,S_n) \ \mathrm{d}r \nonumber\\
&=\sum_{m=1}^M \pi_m \sqrt{\frac{|S_n|}{|\Sigma_m + S_n|}} \ K(x_n|\mu_m,\Sigma_m + S_n) \nonumber\\
&=\sum_{m=1}^M P_m(x_n|S_n)\ . \nonumber
\end{align}

Since the density model is a GMM, we can compute analytically the integral of the product between the balloon kernel and each mixture component. Therefore, multivariate data-adaptive regularizing kernels have the following simple closed form:
\begin{align}
R_n = \sum_{m=1}^M \frac{P_m(x_n|S_n)}{P(x_n|S_n)} \left[\Sigma_{m|S_n} + (x_n-\mu_{m|S_n})(x_n-\mu_{m|S_n})^\top\right] \nonumber
\end{align}
with the parameters of all products with the balloon kernel:
\begin{align}
\Sigma_{m|S_n} &= \left(\Sigma_m^{-1} + S_n^{-1}\right)^{-1} , \nonumber\\
\mu_{m|S_n} &= \Sigma_{m|S_n} \left(\Sigma_m^{-1} \mu_m + S_n^{-1} x_n\right). \nonumber
\end{align}

We apply a balloon estimator independently at every input data samples $x_n$, giving us the convex optimization problems
\begin{align}
\sigma_n=\operatorname*{argmin}_\sigma P(x_n|R_n) = P \quad\text{for each}\quad n \in \left[1,\dots,N\right]. \nonumber
\end{align}
In practice, only a few (or even a single) steps of a fixed point iteration may be used for updating $\sigma_n$ since the objective increases monotonously with the variance. This gives the following sequence of multiplicative updates starting with $\sigma_n \leftarrow 1$: 
\begin{align}
\sigma^2_n \leftarrow \frac{P}{P(x_n|R_n)} \ \sigma^2_n \quad\text{for each}\quad n \in \left[1,\dots,N\right]. \nonumber
\end{align}
We quit the loop whenever $(P(x_n|R_n) - P)^2 < (P \times 0.01)^2$.

This solution is a continuous variant of $K$-nearest neighbors (KNN), where the hard disc indicator function is replaced by a soft multivariate Gaussian kernel that steers to trends in the data and the count $K<N$ is replaced by the probability $P < 1$.

\subsection{Regularized expectation-maximization}
The E-step computes a partition of unity for each "inflated" point sample $x_n$. Sample's positions are exact and the regularized mixture model represents the continuous density including an estimate of missing data samples. Thus, the E-step is simply
\begin{align}
P_{m,n}&\propto \pi_m \ \mathcal{N}\left(x_n|\mu_m,\Sigma_m\right). \nonumber
\end{align}
The M-step updates prior probabilities, means and matrices by using the the regularizing matrices $R_n$ from the balloon estimator as a prior for the variances-covariance matrices: 
\begin{align}
\pi_m&= \frac{1}{N} \sum_{n=1}^N P_{m,n}\ ,\quad \mu_m=\frac{1}{N\pi_m} \sum_{n=1}^N P_{m,n} \ x_n \ , \nonumber\\
\Sigma_m&= \frac{1}{N\pi_m} \sum_{n=1}^N P_{m,n} \left[(x_n-\mu_m) (x_n-\mu_m)^\top + R_{n|m}\right]. \nonumber
\end{align}
with the regularizing additive matrices
\begin{align}
R_{n|m} &= R_n - \left[\Sigma_{m|R_n} + (x_n - \mu_{m|R_n}) (x_n - \mu_{m|R_n})^\top\right]. \nonumber
\end{align}



\section{Experiments}
We have drawn $N = 64$ random samples in a square and iterated 1000 times, until almost sure convergence. Figure~\ref{fig:experiments} shows two results with moderate and strong smoothing  prior probabilities, comparing data overfitting with a fair adaptive smoothing. The adaptive KDE density is simply using the augmented data with multivariate regularizing kernels at sample's positions. Oriented kernels with elongated elliptic footprints have smaller determinants compared to their corresponding balloons. Thus, they apply lighter constraints on parametric estimations.

Experiments demonstrate that even if the balloon estimator computes a single scalar variance per data sample, the adapted regularizing kernels steer to the shape of local data clusters. The EM approach yields much fewer representative mixture components without noticeable loss of quality. The solutions converged to only $45$ and $21$ effective components for $P=1/64$ and $P=1/32$, respectively. Results are comparable to model selection through variational Bayes inference \cite{Tzikas2008}, but without using the (\emph{a priori}) conjugate priors assumption.

\section{Conclusion}
This work introduces a regularized expectation-maximization method for estimating mixture density models of arbitrary size given an incomplete set of point samples. A probability parameter drives the complexity of the solution from maximum-likelihood with one component per sample to ordinary least squares with a single component.
We estimate full-covariance kernels for anisotropic local regularization.
Sparse GMMs are subsuming the data in an effective way and densities are similar to adaptive KDE.
Ongoing work tackle image approximation and restoration using such a regularized continuous model \cite{Schretter2018}.


\end{document}